# On Bayesian Exponentially Embedded Family for Model Order Selection

Zhenghan Zhu, *Member, IEEE*, and Steven Kay, *Life Fellow, IEEE*

*Abstract*—In this paper, we derive a Bayesian model order selection rule by using the exponentially embedded family (EEF) method, termed Bayesian EEF. It shows that the Bayesian EEF can use vague proper priors and improper noninformative priors to be objective in the elicitation of parameter priors. Moreover, the penalty term of the rule is shown to be the sum of half of the parameter dimension and the estimated mutual information between the parameter and observed data. This helps to reveal the EEF mechanism in selecting model orders and may provide new insights into the open problem of choosing an optimal penalty term for model order selection from information-theoretic viewpoints. The Bayesian EEF that uses a *g*-prior is derived for the linear model. The Bayesian EEF is extended for nonlinear models by using Jeffreys' prior. Interestingly, it coincides with the EEF rule derived by a frequentist strategy. This shows another relationship between the frequentist and Bayesian philosophies for model selection.

## I. INTRODUCTION

MODEL order selection is an important problem of active research in signal processing. In model-based signal processing, one often needs to estimate both the number of unknown parameters and their values such as determining the order of autoregressive model [1], and the number of sinusoidal components in a noisy signal [2]. For instance, the determination of the number of sources in array signal processing [3] is essentially a model order selection problem. Overestimating the order fits the noise in the data; underestimating the order, on the other hand, fails to describe the data precisely [3]. Hence, a good model order selection rule is crucial for signal processing applications.

As a multiple hypotheses testing problem, model order selection lacks an optimal solution [4]. The generalized likelihood ratio test (GLRT) always favors the most complex model [5]. A typical model order selection algorithm introduces a penalty term to the GLRT, and it is the penalty term that makes one model order selection rule different from another. A model order selection rule derived from a Bayesian viewpoint typically tries to strike a balance between goodness of fit and model complexity [6].

Some leading algorithms, both frequentist and Bayesian, in the literature [2], [7] are Akaike's information criterion (AIC) [8], the minimum description length (MDL) [9], Bayesian information criterion (BIC) [10] and maximum a posteriori (MAP) [4]. For example, AIC and BIC rules are respectively

$$\ln p(\mathbf{x}|\hat{\boldsymbol{\theta}}) - k; \quad \text{AIC}$$

$$\ln p(\mathbf{x}|\hat{\boldsymbol{\theta}}) - \frac{k}{2}\ln N; \quad \text{BIC}$$

where $\boldsymbol{\theta}$ is the model unknown parameter vector, $\ln p(\mathbf{x}|\hat{\boldsymbol{\theta}})$ is the maximum log-likelihood under a certain model, $k$ is the dimension of the model parameters, $N$ is the data record length. As seen the AIC penalty is $k$, the dimension of the unknown parameter. And BIC has a penalty $\frac{k}{2}\ln N$ which depends on the parameter dimension and data length.

As an alternative, an EEF model order selection rule derived from a frequentist viewpoint is introduced in [11]. It proves effective in model order selection and enjoys many great properties. It is consistent, i.e., the probability of correctly choosing the order goes to one as the signal-to-noise ratio (SNR) increases [12]. Its performance is superior to others in several situations including low SNR regime [3], [11]. It also attains optimality in a minimax sense [13]. EEF has been used to determine the source number in array processing [3], and to determine the order of AR processing [11]. It has also been successfully applied to many related areas such as classification and sensor fusion and shown great performances [14]–[16]. Fundamentally different from [11], we derive in this paper the EEF rule from a Bayesian viewpoint, termed the Bayesian EEF, as a novel Bayesian model order selection rule. The key difference lies in the philosophies of viewing the unknown parameters. The unknown model parameters are treated as deterministic in [11], but random variable in this paper. Using Bayesian strategies allows us the possibilities to investigate the EEF mechanism in a new framework and from new viewpoints such as information theory and leads to the main contributions of this paper:

- A new Bayesian model order selection method, Bayesian EEF, is derived. It is proved that the Bayesian EEF can use both vague proper prior and improper non-informative prior for unknown parameters without having the criticism of the Lindley's paradox or the Information paradox.
- An intuitive justification is given in interpreting the Bayesian EEF penalty term. The penalty term is a sum of half the model parameter dimension and the estimated mutual information between model parameters and observed data. This not only helps to reveal the EEF mechanism in model order selection but also sheds lights on the open problem of choosing a good penalty term in model order selection.
- It also shows that the Bayesian EEF using Jeffrey's prior coincides with the EEF derived from a frequentist viewpoint. This is another case of the interesting interaction between the frequentist and Bayesian philosophies and may provide useful insights into the discussion on the difference between the two.

The paper is organized as follows. In Section II we derive the Bayesian EEF order selection rule that uses a vague proper prior for linear model and discuss some desirable properties of the Bayesian EEF. In Section III we justify the Bayesian EEF penalty term. In Section IV we derive the Bayesian EEF via improper non-informative prior, Jeffreys' prior and discuss its interaction with frequentist EEF. An application of the derived EEF rule, estimating the degree of noncircularity of complex-valued random vectors, is presented in Section V. Finally, some conclusions are given in Section VI.

## II. BAYESIAN EEF RULE FOR MODEL SELECTION VIA VAGUE PROPER PRIOR

Suppose there are $M$ candidate models, $\mathcal{M}_0, \mathcal{M}_1, \ldots, \mathcal{M}_{M-1}$, where $\mathcal{M}_0$ is a null/reference model which has no unknown parameters and the model $\mathcal{M}_i$ (for $i = 1, \ldots, M-1$) has an unknown parameter vector $\boldsymbol{\theta}_i$ of dimension $k_i \times 1$. The probability density functions (PDF) of the observed data $\mathbf{x}$ of dimension $N \times 1$ for model $\mathcal{M}_i$ is denoted as $p_i(\mathbf{x})$. From the frequentist viewpoint, the unknown parameters are deterministic. The EEF model order selection rule proposed in [11] adopts this assumption and hence is termed *frequentist* EEF in this paper. On the other hand, a Bayesian model order selection method views the parameter vectors as random. The Bayesian EEF adopts this philosophy. If we know the the model parameter priors, we can compare marginal PDFs of $\mathbf{x}$ of different models or use a MAP rule to choose a model order. But in practice no prior information is available and the first question that arises for a Bayesian model order selection method is the specification of the prior distributions for the unknown parameter vector $\boldsymbol{\theta}_i$. Which prior to choose is a controversial and difficult task [17]. Ideally we want to use a prior with minimal influence on the Bayesian inference. Improper non-informative priors such as uniform distribution and vague prior distributions (a proper prior with large spread) seem to be natural choices because they are objective in that they do not favor one parameter value over another. However, they can, unfortunately, lead to non-sensible answers when used in some Bayesian model selection methods. As shown later Bayesian EEF, on the other hand, can employ these two types of priors and still produce good results. This is a desirable property for a Bayesian model order selection algorithm. In this section, we derive the Bayesian EEF by assigning vague proper priors to unknown parameters. The resultant EEF is called the *reduced* Bayesian EEF.

Zellner's g-prior is widely used in Bayesian inference because of its conjugacy and computational efficiency in computing the marginal likelihoods and its simple, understandable interpretation [18], [19] and has been used to construct various versions of the existing model order selection methods [20], [21]. Loosely speaking, the g-prior places less prior distribution mass in areas of the parameter space where the data is expected to be more informative about the unknown parameters. The vague proper prior adopted herein is constructed by letting the hyperparameter $g$ of a g-prior goes to infinity and hence produce a "flat" and "non-informative" prior. Assume we want to choose a model from the following linear model candidates

$$\mathcal{M}_i : \mathbf{x} = \mathbf{H}_i \boldsymbol{\theta}_i + \mathbf{w}, \ i = 1, \ldots, M-1.$$

where $\boldsymbol{\theta}_i$ is a $k_i \times 1$ unknown parameter vector, $\mathbf{H}_i$ is a $N \times k_i$ design matrix, and $\mathbf{w} \sim \mathcal{N}(\mathbf{0}, \sigma^2 \mathbf{I})$ is additive noise with $\mathbf{I}$ being a $N \times N$ identity matrix. The noise variance $\sigma^2$ is assumed known. For an unknown noise variance, [22] derives an EEF rule from a frequentist viewpoint. We will extend the Bayesian EEF to the case of unknown $\sigma^2$ in our future work. There is also a null model $\mathcal{M}_0 : \mathbf{x} = \mathbf{w}$ which does not contain unknown parameters. Without loss of generality, we assume that $k_i \leq k_j$ for $i \leq j$.

For $i = 1, \ldots, M-1$, first assign $\boldsymbol{\theta}_i$ a vague proper prior, $\pi_i(\boldsymbol{\theta}_i)$, which is a g-prior with an infinite hypeparameter $g_i$ [19] as

$$\pi_i(\boldsymbol{\theta}_i) = \mathcal{N}\left(\mathbf{0}, g_i \sigma^2 (\mathbf{H}_i^T \mathbf{H}_i)^{-1}\right) \text{ and } g_i \to \infty.$$

The marginal PDF $p_i(\mathbf{x})$ under the $\mathcal{M}_i$ model is then

$$\begin{aligned} p_i(\mathbf{x}) &= \int p_i(\mathbf{x}|\boldsymbol{\theta}_i) \pi_i(\boldsymbol{\theta}_i) d\boldsymbol{\theta}_i \\ &= \mathcal{N}(\mathbf{0}, \sigma^2 \mathbf{I} + g_i \sigma^2 \mathbf{P}_i) \\ &= \mathcal{N}(\mathbf{0}, \mathbf{C}_i) \end{aligned} \quad (1)$$

where $p_i(\mathbf{x}|\boldsymbol{\theta}_i) = \mathcal{N}(\mathbf{H}_i \boldsymbol{\theta}_i, \sigma^2 \mathbf{I})$ is the conditional PDF of $\mathbf{x}$ on $\boldsymbol{\theta}_i$ under model $\mathcal{M}_i$, $\mathbf{P}_i = \mathbf{H}_i (\mathbf{H}_i^T \mathbf{H}_i)^{-1} \mathbf{H}_i^T$ and the covariance matrix $\mathbf{C}_i = \sigma^2 \mathbf{I} + g_i \sigma^2 \mathbf{P}_i$. The PDF of $\mathbf{x}$ under the null model is

$$p_0(\mathbf{x}) = \mathcal{N}(\mathbf{0}, \sigma^2 \mathbf{I}) = \mathcal{N}(\mathbf{0}, \mathbf{C}_0), \quad (2)$$

where $\mathbf{C}_0 = \sigma^2 \mathbf{I}$ is the covariance matrix of $p_0(\mathbf{x})$. Then for each $p_i(\mathbf{x})$, $i = 1, \ldots, M-1$, we can construct a new PDF, $p(\mathbf{x}; \eta_i)$ by exponentially embedding $p_i(\mathbf{x})$ with $p_0(\mathbf{x})$, which is parameterized by an embedding parameter $\eta_i$:

$$\begin{aligned} p(\mathbf{x}; \eta_i) &= \frac{p_i^{\eta_i}(\mathbf{x}) p_0^{1-\eta_i}(\mathbf{x})}{\int p_i^{\eta_i}(\mathbf{x}) p_0^{1-\eta_i}(\mathbf{x}) d\mathbf{x}} \\ &= \exp(\eta_i T_i(\mathbf{x}) - K_0(\eta_i) + f_c(\mathbf{x})) \end{aligned} \quad (3)$$

with

sufficient statistic: $T_i(\mathbf{x}) = \ln \dfrac{p_i(\mathbf{x})}{p_0(\mathbf{x})}$

natural parameter: $0 \leq \eta_i \leq 1$

log-normalizer: $K_0(\eta_i) = \ln \int p_i^{\eta_i}(\mathbf{x}) p_0^{1-\eta_i}(\mathbf{x}) d\mathbf{x}$

$\qquad\qquad\qquad\quad = \ln E_0\left(e^{\eta_i T_i(\mathbf{x})}\right)$

carrier density: $f_c(\mathbf{x}) = \ln p_0(\mathbf{x})$

As shown the resulting PDF belongs to the exponential family and consequently inherits a multitude of mathematical and practical properties of the family. Note that the PDFs $p_i(\mathbf{x})$ and $p_0(\mathbf{x})$ are not necessarily members of the exponential family. The statistic $T_i(\mathbf{x})$ is a minimal and complete sufficient statistic for $\eta_i$; its moments can be easily found and $K_0(\eta_i)$ is a convex function. The new PDF $p(\mathbf{x}; \eta_i)$ is called the Bayesian EEF for the model $\mathcal{M}_i$ in that we employ both Bayesian philosophies and exponentially embedding to construct it. From the information-geometric viewpoints, the log-Bayesian EEF $\ln p(\mathbf{x}; \eta_i)$ can be viewed as a point on the geodesic that connects $\ln p_i(\mathbf{x})$ and $\ln p_0(\mathbf{x})$ [11], [23]. From (3), the Bayesian EEF $p(\mathbf{x}; \eta_i)$ reduces to $p_0(\mathbf{x})$ when $\eta_i = 0$ and $p_i(\mathbf{x})$ when $\eta_i = 1$.

Plugging $p_i(\mathbf{x})$ of (1) and $p_0(\mathbf{x})$ of (2) into (3) produces the reduced Bayesian EEF $p(\mathbf{x}; \eta_i)$ for the linear model as follows.

$$p(\mathbf{x}; \eta_i) = \frac{p_i^{\eta_i}(\mathbf{x}) p_0^{1-\eta_i}(\mathbf{x})}{\exp(K_0(\eta_i))}$$

$$= \frac{1}{\exp(K_0(\eta_i))} \left[ \frac{1}{\sqrt{|2\pi \mathbf{C}_i|}} \exp\left(-\frac{1}{2}\mathbf{x}^T \mathbf{C}_i^{-1} \mathbf{x}\right) \right]^{\eta_i}$$

$$\cdot \left[ \frac{1}{\sqrt{|2\pi \mathbf{C}_0|}} \exp\left(-\frac{1}{2}\mathbf{x}^T \mathbf{C}_0^{-1} \mathbf{x}\right) \right]^{1-\eta_i}$$

$$= c_1 \exp\left[ -\frac{1}{2}\mathbf{x}^T \underbrace{(\eta_i \mathbf{C}_i^{-1} + (1-\eta_i) \mathbf{C}_0^{-1})}_{\mathbf{C}_{\eta_i}^{-1}} \mathbf{x} \right]$$

where $c_1$ is a constant normalization term and $\mathbf{C}_{\eta_i} = \left( \eta_i \mathbf{C}_i^{-1} + (1-\eta_i) \mathbf{C}_0^{-1} \right)^{-1}$. It shows that the constructed EEF is also a zero mean normal distribution with a covariance matrix $\mathbf{C}_{\eta_i}$ depending on $\eta_i$. Explicitly,

$$\mathbf{C}_{\eta_i} = \left[ \eta_i (\sigma^2 \mathbf{I} + g_i \sigma^2 \mathbf{P}_i)^{-1} + (1-\eta_i)(\sigma^2 \mathbf{I})^{-1} \right]^{-1}$$

$$= \sigma^2 \left[ \eta_i \left( \mathbf{I} - \frac{g_i}{g_i+1} \mathbf{P}_i \right) + (1-\eta_i) \mathbf{I} \right]^{-1}$$

$$= \sigma^2 \left( \mathbf{I} - \frac{\eta_i g_i}{g_i+1} \mathbf{P}_i \right)^{-1}$$

$$= \sigma^2 \left( \mathbf{I} + \frac{\eta_i}{1-\eta_i + \frac{1}{g_i}} \mathbf{P}_i \right)$$

$$\to \sigma^2 \mathbf{I} + \frac{\eta_i}{1-\eta_i} \sigma^2 \mathbf{P}_i \text{ as } g_i \to \infty$$

So the reduced Bayesian EEF for $\mathcal{M}_i$ is

$$p(\mathbf{x}; \eta_i) = \mathcal{N}\left( \mathbf{0}, \sigma^2 \mathbf{I} + \frac{\eta_i}{1-\eta_i} \sigma^2 \mathbf{P}_i \right). \quad (4)$$

Then a model order selection algorithm based on the Bayesian EEF in (4) consists of two steps.

- Step1: Find the MLE of $\eta_i$, $0 \le \hat{\eta}_i \le 1$, which maximizes $p(\mathbf{x}; \eta_i)$;
  For the linear model from (4) we have

$$\hat{\eta}_i = \begin{cases} 0 & \text{if } \mathbf{x}^T \mathbf{P}_i \mathbf{x} < k_i \sigma^2 \\ \frac{\mathbf{x}^T \mathbf{P}_i \mathbf{x} - k_i \sigma^2}{\mathbf{x}^T \mathbf{P}_i \mathbf{x}} & \text{otherwise} \end{cases} \quad (5)$$

  where $k_i$ is the dimension of $\boldsymbol{\theta}_i$.
- Step2: Compare the values of the $M-1$ maximized EEF $p(\mathbf{x}; \hat{\eta}_i)$ or equivalently the log-likelihood ratio (LLR) $\ln \frac{p(\mathbf{x};\hat{\eta}_i)}{p_0(\mathbf{x})}$ and choose the model which is associated with the maximum value.

For the linear model, plugging $\hat{\eta}_i$ into (4) produces the maximized LLR

$$\ln \frac{p(\mathbf{x}; \hat{\eta}_i)}{p_0(\mathbf{x})} = \left( \frac{\mathbf{x}^T \mathbf{P}_i \mathbf{x}}{2\sigma^2} - \frac{k_i}{2} - \frac{k_i}{2} \ln \frac{\frac{\mathbf{x}^T \mathbf{P}_i \mathbf{x}}{\sigma^2}}{k_i} \right)$$

$$\cdot u\left( \frac{\mathbf{x}^T \mathbf{P}_i \mathbf{x}}{2\sigma^2} - \frac{k_i}{2} \right).$$

where $u(\cdot)$ is a unit step function. In fact, the term $\frac{\mathbf{x}^T \mathbf{P}_i \mathbf{x}}{2\sigma^2}$ is the maximized LRT of the conditional PDF $p_i(\mathbf{x}|\hat{\boldsymbol{\theta}}_i)$ and $p_0(\mathbf{x})$, termed as $l_{G_i}$:

$$l_{G_i} = \ln \frac{\max_{\boldsymbol{\theta}_i} p_i(\mathbf{x}|\boldsymbol{\theta}_i)}{p_0(\mathbf{x})}$$

$$= \ln \frac{p_i(\mathbf{x}|\hat{\boldsymbol{\theta}}_i)}{p_0(\mathbf{x})} \text{ with } \hat{\boldsymbol{\theta}}_i = (\mathbf{H}_i^T \mathbf{H}_i)^{-1} \mathbf{H}_i^T \mathbf{x}$$

$$= \frac{\mathbf{x}^T \mathbf{P}_i \mathbf{x}}{2\sigma^2}$$

where the MLE $\hat{\boldsymbol{\theta}}_i$ is the value of $\boldsymbol{\theta}_i$ that maximizes $p_i(\mathbf{x}|\boldsymbol{\theta}_i)$ or explicitly,

$$p_i(\mathbf{x}|\boldsymbol{\theta}_i) = \frac{\exp\left(-\frac{1}{2\sigma^2}(\mathbf{x} - \mathbf{H}_i \boldsymbol{\theta}_i)^T (\mathbf{x} - \mathbf{H}_i \boldsymbol{\theta}_i)\right)}{\sqrt{|2\pi\sigma^2 \mathbf{I}|}}$$

In summary, we can write the linear model Bayesian EEF as

$$\ln \frac{p(\mathbf{x}; \hat{\eta}_i)}{p_0(\mathbf{x})} = \left( l_{G_i} - \frac{k_i}{2} - \frac{k_i}{2} \ln \frac{l_{G_i}}{k_i/2} \right) u\left( l_{G_i} - \frac{k_i}{2} \right). \quad (6)$$

### A. Rationale of Bayesian EEF Model Order Selection Algorithm

The rationale for Bayesian EEF model order selection algorithm is as follows. When $\eta_i$ is chosen as its MLE $\hat{\eta}_i$,

$$\frac{\partial \ln p(\mathbf{x}; \eta_i)}{\partial \eta_i} = T_i(\mathbf{x}) - K_0'(\eta_i) = 0$$

follows from (3). That is $T_i(\mathbf{x}) = K_0'(\eta_i)$ evaluated at $\eta_i = \hat{\eta}_i$. Moreover, it holds in general $\int p(\mathbf{x}; \eta_i) T_i(\mathbf{x}) d\mathbf{x} = K_0'(\eta_i)$ for the exponential family [11]. Therefore

$$\left[ \int p(\mathbf{x}; \eta_i) T_i(\mathbf{x}) d\mathbf{x} \right]\bigg|_{\eta_i \to \hat{\eta}_i} = T_i(\mathbf{x}) \quad (7)$$

And consequently we have

$$\text{KL}(p(\mathbf{x}; \hat{\eta}_i) || p_0(\mathbf{x}))$$

$$= \int p(\mathbf{x}; \hat{\eta}_i) \ln \frac{p(\mathbf{x}; \hat{\eta}_i)}{p_0(\mathbf{x})} d\mathbf{x}$$

$$= \int p(\mathbf{x}; \hat{\eta}_i) [\hat{\eta}_i T_i(\mathbf{x}) - K_0(\hat{\eta}_i)] d\mathbf{x}$$

$$= \hat{\eta}_i T_i(\mathbf{x}) - K_0(\hat{\eta}_i)$$

$$= \ln \frac{p(\mathbf{x}; \hat{\eta}_i)}{p_0(\mathbf{x})} \quad (8)$$

where $\text{KL}(\cdot || \cdot)$ denotes Kullback Libler divergence (KLD).

Moreover, a Pythagorean-like relationship holds asymptotically for large data record among KLD quantities for EEF [11]

$$\mathrm{KL}(p_t(\mathbf{x})||p(\mathbf{x};\hat{\eta}_i)) = \mathrm{KL}(p_t(\mathbf{x})||p_0(\mathbf{x}))$$
$$- \mathrm{KL}(p(\mathbf{x};\hat{\eta}_i)||p_0(\mathbf{x})),$$

where $p_t(\mathbf{x})$ denotes the true PDF of the data, which is unknown but fixed. The distance $\mathrm{KL}(p_t(\mathbf{x})||p_0(\mathbf{x}))$ is fixed, hence the model that maximizes the distance $\mathrm{KL}(p(\mathbf{x};\hat{\eta}_i)||p_0(\mathbf{x}))$ or equivalently $\ln \frac{p(\mathbf{x};\hat{\eta}_i)}{p_0(\mathbf{x})}$, among all models has the minimum $\mathrm{KL}(p_t(\mathbf{x})||p(\mathbf{x};\hat{\eta}_i))$-the "distance" from the true PDF $p_t(\mathbf{x})$. This is the reason why the Bayesian EEF model selection rule chooses the model associated with the maximum of the maximized EEF's.

*B. Discussion on Paradoxes*

The EEF model order selection algorithm has many desirable properties such as consistency [3] and better performances than many other algorithms in the low signal-to-noise ratio regime [11]. In addition to these properties, we now show that the newly derived Bayesian EEF has additional desirable properties-it does not have *Lindley's paradox* nor the *Information paradox*. On the contrary, many other Bayesian model selection methods based on marginal Bayes factor (BF) may suffer from these paradoxes [18]. *Lindley's paradox* can be understood as: "large spread of the prior induced by the non-informative choice of hyper-parameter has the unintended consequence of forcing the BF to favor the null model, the smallest model, regardless of the information in the data [18]". As shown in (6), the reduced Bayesian EEF does not necessarily favor the null model even if we let the hyper-parameter $g_i \to \infty$. This indicates that the reduced Bayesian EEF rule has no "Lindley's paradox". The *Information paradox* is "a paradox related to the limiting behavior of the BF. The BF yields a constant even when there is an infinite amount of information supporting to choose a model [18]." For instance, the linear model BF resulted from assigning the parameter $\boldsymbol{\theta}_i$ a g-prior with a certain $g_i$ is [18]

$$BF(\mathcal{M}_i : \mathcal{M}_0) = \frac{(1+g_i)^{(N-1-k_i)/2}}{(1+g_i(1-R_r^2))^{(N-1)/2}}$$

where $R_r^2$ is the ordinary coefficient of determination of the regression model $\mathcal{M}_i$, $R_r^2 = \frac{\mathbf{x}^T \mathbf{H}_i (\mathbf{H}_i^T \mathbf{H}_i)^{-1} \mathbf{H}_i \mathbf{x}}{\mathbf{x}^T \mathbf{x}}$, which measures how well the data $\mathbf{x}$ fits the linear regression [24]. When there is overwhelming information supporting to choose $\mathcal{M}_i$ instead of $\mathcal{M}_0$, $R_r^2 \to 1$; however, the BF yields a constant $(1+g_i)^{(N-1-k_i)/2}$ instead of infinity. This information limiting behavior is called the information paradox. When $R_r^2 \to 1$ or equivalently $\mathbf{x}^T \mathbf{P}_i \mathbf{x} \gg k_i \sigma^2$ we have $\hat{\eta}_i \to 1$ from (5). In this case, the reduced Bayesian EEF $\ln \frac{p(\mathbf{x};\hat{\eta}_i)}{p_0(\mathbf{x})}$ in (6) also goes to infinity. This shows that the Bayesian EEF has no information limiting behavior and hence no *Information paradox*. These two nice properties of the Bayesian EEF model selection rule are due to its mechanism of choosing the value of $\eta_i$. It uses the MLE $\hat{\eta}_i$ which is dependent on data. This is similar to the mechanism why empirical Bayesian methods [25]–[27] also enjoy the nice properties of not having the Lindley paradox or the information paradox: replacing unknown hyperparameters with their data-dependent estimates [21]. However, note that the EEF method is fundamentally different from empirical Bayesian methods. Bayesian EEF method embeds the candidate model PDF which is assigned a vague proper prior with a reference PDF, estimates the embedding coefficient, and enjoys the properties of the exponential family. Empirical Bayesian methods estimate the prior from the data.

III. THE PENALTY TERM OF REDUCED BAYESIAN EEF

The penalty term is the key term for a model order selection rule. Its function is to penalize the maximum log-likelihood with a measure of model complexity so that the model order selection rule can strike a tradeoff between goodness-of-fit and model complexity. Some useful discussion on its design is given in [28]. In light of the general relationship KLD = SNR-MI [29], the reduced Bayesian EEF penalty term is found to possess a very intuitive and enlightening interpretation. This not only helps further understanding EEF's mechanism in model selection but also provides new insights into the problem of choosing a good penalty term for model selection. As shown next, the EEF penalty term can be viewed as the sum of a term proportional to the parameter dimension, $\frac{k_i}{2}$, and estimated mutual information between the parameter and received data, $\frac{k_i}{2} \ln \frac{2l_{G_i}}{k_i}$.

When assigning the unknown parameter $\boldsymbol{\theta}_i$ a prior that depends upon the embedding parameter $\eta_i$:

$$\pi'(\boldsymbol{\theta}_i; \eta_i) = \mathcal{N}\left(\mathbf{0}, \frac{\eta_i}{1-\eta_i}\sigma^2(\mathbf{H}_i^T \mathbf{H}_i)^{-1}\right),$$

the marginal PDF for model $\mathcal{M}_i$ becomes the reduced Bayesian EEF in (4)

$$p_i(\mathbf{x}) = \int p_i(\mathbf{x}|\boldsymbol{\theta}_i)\pi'(\boldsymbol{\theta}_i;\eta_i)d\boldsymbol{\theta}_i$$
$$= \mathcal{N}\left(\mathbf{0}, \sigma^2 \mathbf{I} + \frac{\eta_i}{1-\eta_i}\sigma^2 \mathbf{P}_i\right)$$
$$= p(\mathbf{x};\eta_i).$$

Note that this new $p_i(\mathbf{x})$ is in fact parameterized by $\eta_i$ because $\pi'(\boldsymbol{\theta}_i;\eta_i)$ depends upon $\eta_i$. To strengthen this point, we denote $p_i(\mathbf{x})$ as $p_{\eta_i}(\mathbf{x})$, then $p_{\eta_i}(\mathbf{x}) = p(\mathbf{x};\eta_i)$. With the relationship of (8) and the decomposition KLD = SNR − MI established in [29] (see also [30] for some illustrative examples of this decomposition), we have, for $\eta_i = \hat{\eta}_i$,

$$\ln \frac{p(\mathbf{x};\eta_i)}{p_0(\mathbf{x})} \approx \mathrm{KL}(p(\mathbf{x};\eta_i)||p_0(\mathbf{x}))$$
$$= \mathrm{KL}(p_{\eta_i}(\mathbf{x})||p_0(\mathbf{x}))$$
$$= \int p_{\eta_i}(\mathbf{x}) \ln \frac{p_{\eta_i}(\mathbf{x})}{p_0(\mathbf{x})} d\mathbf{x}$$
$$= \underbrace{\int\int p_{\eta_i}(\mathbf{x},\boldsymbol{\theta}_i) \ln \frac{p_{\eta_i}(\mathbf{x}|\boldsymbol{\theta}_i)}{p_0(\mathbf{x})} d\boldsymbol{\theta}_i d\mathbf{x}}_{\widehat{\mathrm{SNR}}}$$
$$- \underbrace{\int\int p_{\eta_i}(\mathbf{x},\boldsymbol{\theta}_i) \ln \frac{p_{\eta_i}(\mathbf{x}|\boldsymbol{\theta}_i)}{p_{\eta_i}(\mathbf{x})} d\mathbf{x} d\boldsymbol{\theta}_i \quad (9)}_{\widehat{\mathrm{MI}}}$$

where $p_{\eta_i}(\mathbf{x},\boldsymbol{\theta}_i)$ denotes the joint PDF of $\mathbf{x}$ and $\boldsymbol{\theta}_i$ and $p_{\eta_i}(\mathbf{x}|\boldsymbol{\theta}_i) = \mathcal{N}(\mathbf{H}_i\boldsymbol{\theta}_i, \sigma^2 \mathbf{I})$ is the conditional PDF. The eqn (19) suggests that the reduced EEF can be decomposed into two terms. As shown next, the first term is an estimated SNR, denoted as $\widehat{\text{SNR}}$ and the second term is an estimated MI between parameter $\boldsymbol{\theta}_i$ and data $\mathbf{x}$, denoted as $\widehat{\text{MI}}$. They are estimated terms in the sense that $\eta_i$ is replaced by its MLE $\hat{\eta}_i$. We next elaborate each term.

### A. The Estimated SNR Term

For $\eta_i = \hat{\eta}_i$, we have proved in Appendix A.1 that

$$\widehat{\text{SNR}} = \int \int p_{\eta_i}(\mathbf{x}, \boldsymbol{\theta}_i) \ln \frac{p_{\eta_i}(\mathbf{x}|\boldsymbol{\theta}_i)}{p_0(\mathbf{x})} d\boldsymbol{\theta}_i \, d\mathbf{x} \quad (10)$$

$$= \int_{\boldsymbol{\theta}_i} \pi'(\boldsymbol{\theta}_i) \left( \frac{1}{2} \frac{\boldsymbol{\theta}_i^T \mathbf{H}_i^T \mathbf{H}_i \boldsymbol{\theta}_i}{\sigma^2} \right) d\boldsymbol{\theta}_i \quad (11)$$

$$= l_{G_i} - \frac{k_i}{2} \quad (12)$$

The eqn (11) indicates that the first term is an average ratio of signal energy $\|\mathbf{H}_i \boldsymbol{\theta}_i\|^2$ and the noise power $\sigma^2$, and indeed is a measure of SNR; furthermore by (12) we see that $\widehat{\text{SNR}}$ has introduced a penalty term $k_i/2$, which is proportional to the parameter dimension. In fact, (12) not only holds for linear model but also is approximately valid in general for large data record as proved in Appendix A.2. This shows that the difference between $l_{G_i}$ and the estimated SNR is asymptotically half of the parameter dimension in general.

### B. The Estimated Mutual Information Term

We next consider the second term $\widehat{\text{MI}}$ in the decomposition (9). From the definition of mutual information, we have (see Appendix A.3 for details),

$$\widehat{\text{MI}}$$
$$= \int \int p_{\hat{\eta}_i}(\mathbf{x}, \boldsymbol{\theta}_i) \ln \frac{p_{\hat{\eta}_i}(\mathbf{x}|\boldsymbol{\theta}_i)}{p_{\hat{\eta}_i}(\mathbf{x})} d\mathbf{x} d\boldsymbol{\theta}_i \quad (13)$$

$$= \frac{k_i}{2} \ln \left( \frac{1}{1 - \hat{\eta}_i} \right) \quad (14)$$

$$= \frac{k_i}{2} \ln \left( \frac{\mathbf{x}^T \mathbf{P}_i \mathbf{x}}{k_i \sigma^2} \right) \quad (15)$$

$$= \frac{k_i}{2} \ln \frac{2 l_{G_i}}{k_i} \quad (16)$$

This verifies that the term $\frac{k_i}{2} \ln \frac{2 l_{G_i}}{k_i}$ of (6) is indeed an estimated mutual information between between $\boldsymbol{\theta}_i$ and $\mathbf{x}$. As a measure of the statistical dependence of the parameter and observed data, the estimated MI is a reasonable measure of model complexity. First, the estimated MI can be viewed as averaged KLD distance between the $p_{\eta_i}(\mathbf{x}|\boldsymbol{\theta}_i)$ and $p_{\eta_i}(\mathbf{x})$, see (13), which assesses the "modeling potential" of the conditional distribution. Second, the estimated MI also measures the difference between the prior and posterior distributions of the unknown parameter and thus relates to the "difficulty of estimation" [32]. From (14) we see that for linear model $\widehat{\text{MI}}$ is monotonic with both the parameter dimension $k_i$ and the embedding parameter $\hat{\eta}_i$. As $\hat{\eta}_i$ goes to zero, $\widehat{\text{MI}} \to 0$. This is in agreement with the expectation from (3) in that when $\eta_i \to 0$, the Bayesian EEF $p(\mathbf{x}; \eta_i)$ reduces to the null model PDF $p_0(\mathbf{x})$. When $\hat{\eta}_i$ increases, the resulting Bayesian EEF $p(\mathbf{x}; \eta_i)$ moves closer towards $p_i(\mathbf{x})$ as shown in (3). The estimated MI simultaneously increases to reflect the increasing model complexity.

As shown, the Bayesian EEF penalty term takes into account three levels of model complexity, namely, parameter dimension, the prior of the unknown parameter $\pi'_i(\boldsymbol{\theta}_i)$ and the functional form on how the model is parameterized, the latter two of which contribute to the estimated MI. On the other hand, AIC only accounts for the dimension of unknown parameters $k_i$; BIC takes into consideration the parameter dimension $k_i$ and the number of independently identical distributed (IID) data samples [8], [10] and [33].

### C. An Alternative Interpretation of the Estimated Mutual Information Term

A closer look at the estimated mutual information term in (16) leads to an alternative intuition. Using the approximate relationship of $\widehat{\text{SNR}}$ and $l_{G_i}$ (31) in (16) we have

$$\widehat{\text{MI}} = \frac{k_i}{2} \ln \frac{2 l_{G_i}}{k_i}$$

$$= k_i \underbrace{\left[ \frac{1}{2} \ln \left( 1 + \frac{\widehat{\text{SNR}}}{k_i / 2} \right) \right]}_{\widehat{\text{MI}} \text{ per dim}}$$

The estimated mutual information term is the multiplicative result of parameter dimension $k_i$ and the estimated MI per parameter dimension $\frac{1}{2} \ln(1 + \frac{\widehat{\text{SNR}}}{k_i/2})$. As an example, for the normal linear model we have from (15) that $\widehat{\text{MI}} = \frac{k_i}{2} \ln(\frac{\mathbf{x}^T \mathbf{P}_i \mathbf{x}}{k_i \sigma^2})$ and

$$\mathbf{x}^T \mathbf{P}_i \mathbf{x}$$
$$= \mathbf{x}^T \mathbf{H}_i (\mathbf{H}_i^T \mathbf{H}_i)^{-1} \mathbf{H}_i^T \mathbf{x}$$
$$= \mathbf{x}^T \mathbf{H}_i (\mathbf{H}_i^T \mathbf{H}_i)^{-\frac{1}{2}} (\mathbf{H}_i^T \mathbf{H}_i)^{-\frac{1}{2}} \mathbf{H}_i^T \mathbf{x}$$
$$= \| \underbrace{(\mathbf{H}_i^T \mathbf{H}_i)^{-\frac{1}{2}} \mathbf{H}_i^T \mathbf{x}}_{\mathbf{y}} \|^2$$
$$= \| (\mathbf{H}_i^T \mathbf{H}_i)^{-\frac{1}{2}} \mathbf{H}_i^T (\mathbf{H}_i \boldsymbol{\theta}_i + \mathbf{w}) \|^2$$
$$= \| \underbrace{(\mathbf{H}_i^T \mathbf{H}_i)^{-\frac{1}{2}} \mathbf{H}_i^T \mathbf{H}_i \boldsymbol{\theta}_i}_{\boldsymbol{\theta}'_i} + \underbrace{(\mathbf{H}_i^T \mathbf{H}_i)^{-\frac{1}{2}} \mathbf{H}_i^T \mathbf{w}}_{\mathbf{w}'} \|^2$$

where we have denoted $\boldsymbol{\theta}'_i = (\mathbf{H}_i^T \mathbf{H}_i)^{-\frac{1}{2}} \mathbf{H}_i^T \mathbf{H}_i \boldsymbol{\theta}_i = (\mathbf{H}_i^T \mathbf{H}_i)^{\frac{1}{2}} \boldsymbol{\theta}_i$. It is of dimension $k_i \times 1$ and can be viewed as a signal coordinate vector. Also $\mathbf{w}' = (\mathbf{H}_i^T \mathbf{H}_i)^{-\frac{1}{2}} \mathbf{H}_i^T \mathbf{w}$ is of dimension $k_i \times 1$ and is a noise coordinate vector. Finally we denote $\mathbf{y} = \boldsymbol{\theta}'_i + \mathbf{w}'$, which is of dimension $k_i \times 1$.

With these notations, the estimated MI can be rewritten as

$$\widehat{\text{MI}} = \frac{k_i}{2} \ln \left( \frac{\| \boldsymbol{\theta}'_i + \mathbf{w}' \|^2}{k_i \sigma^2} \right) \quad (17)$$

$$= \frac{k_i}{2} \ln \left( \frac{\frac{1}{k_i} \sum_{j=1}^{k_i} (\theta'_i[j] + w'[j])^2}{\sigma^2} \right) \quad (18)$$

where $\theta_i'[j]$ and $w'[j]$ are the $j$th elements of the vector $\theta_i'$ and $\mathbf{w}'$ respectively.

Furthermore, we have the distributions of $\theta_i'$ and $\mathbf{w}'$ based on the PDFs of $\theta_i$ and $\mathbf{w}$, as

$$\theta_i' \sim \mathcal{N}(\mathbf{0}, \mathbf{C}_{\theta_i'})$$

with

$$\begin{aligned}\mathbf{C}_{\theta_i'} &= (\mathbf{H}_i^T \mathbf{H}_i)^{\frac{1}{2}} \frac{\eta_i}{1-\eta_i} \sigma^2 (\mathbf{H}_i^T \mathbf{H}_i)^{-1} (\mathbf{H}_i^T \mathbf{H}_i)^{\frac{1}{2}} \\ &= \underbrace{\frac{\eta_i}{1-\eta_i}\sigma^2}_{\sigma^2_{\theta_i}} \mathbf{I}_{k_i},\end{aligned}$$

where $\mathbf{I}_{k_i}$ denotes the identity matrix of dimension $k_i$ and we have introduced $\sigma^2_{\theta_i} = \frac{\eta_i}{1-\eta_i}\sigma^2$ to simply the notation. This shows that by using the g-prior on $\theta_i$, the coordinate vector $\theta_i'$ has a scaled identity matrix as its covariance matrix; that is each element of the resulting vector $\theta_i'$ is identically independently distributed (IID). The g-prior equalizes the distribution of each parameter of $\theta_i$.

Similarly, we have the distribution of $\mathbf{w}'$ as

$$\mathbf{w}' \sim \mathcal{N}(\mathbf{0}, \mathbf{C}_{\mathbf{w}'})$$

with

$$\begin{aligned}\mathbf{C}_{\mathbf{w}'} &= (\mathbf{H}_i^T \mathbf{H}_i)^{-\frac{1}{2}} \mathbf{H}_i^T \sigma^2 \mathbf{I}_N \mathbf{H}_i (\mathbf{H}_i^T \mathbf{H}_i)^{-\frac{1}{2}} \\ &= \sigma^2 \mathbf{I}_{k_i}\end{aligned}$$

This shows that $\mathbf{w}'$ still has a zero mean normal distribution with a covariance matrix being $\sigma^2 \mathbf{I}_{k_i}$. Then we have the PDF of $\mathbf{y} = \theta_i' + \mathbf{w}'$, $p(\mathbf{y})$ as

$$\begin{aligned}p(\mathbf{y}) &= \mathcal{N}(\mathbf{0}, \mathbf{C}_{\theta_i'} + \mathbf{C}_{\mathbf{w}'}) \\ &= \mathcal{N}\left(\mathbf{0}, (\sigma^2 + \sigma^2_{\theta_i})\mathbf{I}_{k_i}\right)\end{aligned}$$

In fact the term $\frac{1}{k_i}\sum_{j=1}^{k_i}(\theta_i'[j] + w'[j])^2$ in (18) is the estimate of $\sigma^2 + \sigma^2_{\theta_i}$ and the hence (18) can be expressed alternatively as

$$\widehat{\text{MI}} = k_i \underbrace{\frac{1}{2}\ln\left(\frac{\widehat{\sigma^2 + \sigma^2_{\theta_i}}}{\sigma^2}\right)}_{\widehat{\text{MI per dim}}}$$

The term $\widehat{\text{MI}}$ per dim is the standard estimated mutual information for the case of Gaussian signal in additive Gaussian noise [34] for each signal component/parameter dimension. Since by employing the g-prior each element of the signal $\theta_i'$ is IID, the total estimated MI is simply a multiplication of the $\widehat{\text{MI}}$ per dim and the parameter dimension $k_i$. This provides another intuition on how the estimated MI depends on the parameter dimensions and the mechanism of the g-prior.

It is worth pointing out that the penalty terms of nMDL and gMDL [25] are similar to that of reduced Bayesian EEF. Hence the justification of the Bayesian EEF penalty term may also help to understand better the penalty terms of nMDL and gMDL.

## IV. BAYESIAN EEF VIA JEFFREYS' PRIOR

Jeffreys' prior is another compelling non-informative prior [17] due to its property of invariance to reparameterization. In this section, we use the Jeffreys' prior in Bayesian EEF and derive the *asymptotic* Bayesian EEF. For each model $\mathcal{M}_i$ we assign a Jeffreys' prior $\pi_i(\theta_i)$ to the unknown $\theta_i$. The Jeffreys' prior PDF of $\theta$ is proportional to the square root of the determinant of FIM of $\theta_i$; that is, $\pi_i(\theta_i) \propto \sqrt{|\mathbf{I}(\theta_i)|}$. A motivation for the Jeffreys' prior is that Fisher information $\mathbf{I}(\theta_i)$ is an indicator of the amount of information brought by the model/observations about unknown parameter $\theta_i$. Favoring the values of $\theta_i$ for which $\mathbf{I}(\theta_i)$ is large, is equivalent to minimizing the influence of the prior [17]. By the Laplace approximation we have

$$p_i(\mathbf{x}|\theta_i) \approx p_i(\mathbf{x}|\hat{\theta}_i) e^{-\frac{1}{2}(\theta_i-\hat{\theta}_i)^T \mathbf{I}(\hat{\theta}_i)(\theta_i-\hat{\theta}_i)}.$$

Moreover when assuming that $\pi_i(\theta_i)$ is flat around $\hat{\theta}_i$, which is valid for large data records, we have approximately

$$\begin{aligned}p_i(\mathbf{x}) &= \int_{\theta_i} p_i(\mathbf{x}|\theta_i)\pi_i(\theta_i) d\theta_i \\ &\approx p_i(\mathbf{x}|\hat{\theta}_i)\pi_i(\hat{\theta}_i) \int e^{-\frac{1}{2}(\theta_i-\hat{\theta}_i)^T \mathbf{I}(\hat{\theta}_i)(\theta_i-\hat{\theta}_i)} d\theta_i \\ &= \frac{p_i(\mathbf{x}|\hat{\theta}_i)\pi_i(\hat{\theta}_i)}{(2\pi)^{-\frac{k_i}{2}}\sqrt{|\mathbf{I}(\hat{\theta}_i)|}}\end{aligned}$$

Substituting this approximation into the EEF definition, we have

$$\begin{aligned}&\ln\frac{p(\mathbf{x};\eta_i)}{p_0(\mathbf{x})} \\ &= \eta_i \ln\frac{p_i(\mathbf{x})}{p_0(\mathbf{x})} - K_0(\eta_i) \\ &\approx \eta_i \ln\frac{\frac{p_i(\mathbf{x}|\hat{\theta}_i)\pi_i(\hat{\theta}_i)}{(2\pi)^{-\frac{k_i}{2}}\sqrt{|\mathbf{I}(\hat{\theta}_i)|}}}{p_0(\mathbf{x})} \qquad (19)\\ &\quad - \ln E_0 \exp\left(\eta_i \ln \frac{\frac{p_i(\mathbf{x}|\hat{\theta}_i)\pi_i(\hat{\theta}_i)}{(2\pi)^{-\frac{k_i}{2}}\sqrt{|\mathbf{I}(\hat{\theta}_i)|}}}{p_0(\mathbf{x})}\right) \\ &= \eta_i \ln\frac{p_i(\mathbf{x}|\hat{\theta}_i)}{p_0(\mathbf{x})} - \ln E_0 \exp\left(\eta_i \ln\frac{p_i(\mathbf{x}|\hat{\theta}_i)}{p_0(\mathbf{x})}\right) \qquad (20)\end{aligned}$$

Assigning $\theta_i$ a Jeffreys' prior, the term $\frac{\pi_i(\hat{\theta}_i)}{(2\pi)^{-\frac{k_i}{2}}\sqrt{|\mathbf{I}(\hat{\theta}_i)|}}$ becomes a constant and thus the marginal PDF $p_i(\mathbf{x})$ becomes the multiplication of the maximized conditional PDF $p_i(\mathbf{x}|\hat{\theta}_i)$ with the constant. From the derivation, it shows that by employing EEF mechanism, the resulting Bayesian model selection rule does not suffer from problems when $\int \sqrt{\mathbf{I}(\theta_i)}d\theta_i \to \infty$ as the FIM term is eliminated by the log-normalization term $K_0(\eta_i)$ using the Jeffreys' prior. This is one of many examples showing that the embedded family derives many of its useful properties from the use of the normalization term $K_0(\eta_i)$ [11]. And it is this property that makes the *approximate* Bayesian EEF yield the same result as the frequentist EEF in [11], which can be obtained by using the estimation of the $\eta_i$ [11]

$$\text{BEEF}_i = \left(l_{G_i} - \frac{k_i}{2} - \frac{k_i}{2}\ln\frac{l_{G_i}}{k_i/2}\right)u\left(l_{G_i} - \frac{k_i}{2}\right) \qquad (21)$$

where BEEF$_i$ is the asymptotic Bayesian EEF for model $i$, $l_{G_i} = \ln \frac{p_i(\mathbf{x}|\hat{\boldsymbol{\theta}}_i)}{p_0(\mathbf{x})}$ and $k_i$ is the unknown parameter dimenion under model $\mathcal{M}_i$. The asymptotic Bayesian EEF rule chooses the model that has the maximum BEEF.

As a special case, the reduced Bayesian EEF, approximate Bayesian EEF method and the reduced frequentist EEF all coincide with each other for the normal linear model problem. This coincidence stems from the fact that the FIM for all $\boldsymbol{\theta}_i$ are the same under a certain model $\mathcal{M}_i$ in that $\mathbf{I}(\boldsymbol{\theta}_i) = \frac{\mathbf{H}_i^T \mathbf{H}_i}{\sigma^2}$. In this case the Jeffreys' prior, $\pi(\boldsymbol{\theta}_i) \propto \sqrt{|\mathbf{I}(\boldsymbol{\theta}_i)|}$, becomes an improper uniform distribution, $\pi(\boldsymbol{\theta}_i) = c > 0$, where $c$ is a positive constant. This example also shows that Bayesian EEF can employ improper uniform prior without suffering from integration problems.

## V. AN APPLICATION OF BAYESIAN EEF–ESTIMATING THE DEGREE OF NONCIRCULARITY OF RANDOM COMPLEX-VALUED VECTOR

In this section, we apply our Bayesian EEF model order selection rule to determine the degree of noncircularity of random complex valued vectors and compare its performance with several other leading model selection rules.

A complex-valued signal is often encountered in communication [35], radar [36], [37], sonar [38] and biomedical engineering. Modeling complex data as noncircular often provides better fitting of physical conditions, but requires complicated signal processing algorithms and more computational cost. On the other hand, modeling complex data as circular requires less computational resources but may yield poor representations of true scenarios. Estimating the degree of data's noncircularity is useful. It has been reported that higher resolution of estimation of direction of arrival and independent component analysis can be achieved by properly using the noncirculary models [35]. The estimation of degree is essentially a model order selection problem for a nonlinear model. Hence we apply the approximate Bayesian EEF that uses the Jeffreys' prior to estimate the degree of the noncircularity, which was derived in Section IV.

A complex-valued random vector $\mathbf{x} \in \mathbb{C}^{N \times 1}$ is circular if its probability distribution is invariant to rotation in the complex plane, or equivalently, if its pseudocovariance matrix $\mathbf{P} = E(\mathbf{x}\mathbf{x}^T) = \mathbf{0}$, where $T$ represents transpose. Conversely, it is noncircular if $\mathbf{P} \neq \mathbf{0}$ [35]. Assume we observe $M$ IID data vectors $\mathbf{x}_1, \mathbf{x}_2, \cdots, \mathbf{x}_M$ and each $\mathbf{x}_m$ for $m = 1, 2, \ldots, M$ is a $N \times 1$ complex-valued Gaussian random vector with its mean being zero, $\boldsymbol{\mu}(\mathbf{x}) = \mathbf{0}$. We will denote the received data as $\mathbf{X} = [\mathbf{x}_1 \mathbf{x}_2 \cdots \mathbf{x}_M]$. For a noncircular complex vector $\mathbf{x}$, the conventional covariance matrix $\mathbf{C} = E(\mathbf{x}\mathbf{x}^H)$, where $H$ represents Hermitian, is not sufficient to fully describe its second-order properties. The pseudo-covariance matrix $\mathbf{P} = E(\mathbf{x}\mathbf{x}^T)$ is needed as complementary information. To fully represent a noncircular complex vector's second-order properties, an argumented covariance matrix defined as the covariance matrix of the argumented random vector $\underline{\mathbf{x}} = [\mathbf{x}^T \mathbf{x}^H]^T$,

$$\underline{\mathbf{R}} = E(\underline{\mathbf{x}}\underline{\mathbf{x}}^H) = \begin{bmatrix} \mathbf{C} & \mathbf{P} \\ \mathbf{P}^* & \mathbf{C}^* \end{bmatrix}$$

is often used [35]. With the argumented covariance matrix, the PDF of $\underline{\mathbf{x}}$ can be writen as

$$p(\underline{\mathbf{x}}; \mathbf{C}, \mathbf{P}) = \frac{1}{\pi^N |\underline{\mathbf{R}}|^{\frac{1}{2}}} \exp\left(-\frac{1}{2}\underline{\mathbf{x}}^H \underline{\mathbf{R}}^{-1} \underline{\mathbf{x}}\right)$$

When the data is circular, then $\mathbf{P} = \mathbf{0}$, and the PDF reduces to a regular circularly complex Gaussian distribution.

The circularity coefficients $\lambda_k$'s for $k = 1, 2, \ldots, N$ are defined as singular values of the coherence matrix $\mathbf{C}_h = \mathbf{C}^{-\frac{1}{2}} \mathbf{P} \mathbf{C}^{-\frac{T}{2}}$ [39]. Without loss of generality, let

$$\lambda_1 \geq \lambda_2 \geq \cdots \geq \lambda_N$$

The number of nonzero $\lambda$'s is called the degree of noncircularity. Hence, the problem of estimating the degree of noncircularity is equivalent to choosing one of the following hypotheses/models.

$$\mathcal{H}_1 : \lambda_1 > \lambda_2 = \cdots = \lambda_N = 0, \text{ model } \mathcal{M}_1$$
$$\vdots$$
$$\mathcal{H}_k : \lambda_1 \geq \cdots \geq \lambda_k > \lambda_{k+1} = \cdots = \lambda_N = 0, \text{ model } \mathcal{M}_k$$
$$\vdots$$
$$\mathcal{H}_N : \lambda_1 \geq \lambda_2 \geq \cdots \geq \lambda_N > 0, \text{ model } \mathcal{M}_N \quad (22)$$

Under the model $\mathcal{M}_k$, the degree of noncirculairty is $k$, and it has $d_k$ unknown parameters, written as a vector $\boldsymbol{\theta}_k$.

### A. Asymptotic Bayesian EEF Rule for Noncircularity Degree Estimation

The asymptotic Bayesian EEF rule in (21) for the problem of estimating the degree of noncircularity. The Bayesian EEF rule chooses the $k$ which maximizes the following

$$\text{BEEF}_k = l_k(\mathbf{X}) - d_k \left[\ln\left(\frac{l_k(\mathbf{X})}{d_k}\right) + 1\right] u\left(l_k(\mathbf{X}) - d_k\right),$$

where

$$l_k(\mathbf{X}) = 2\ln \frac{\max p(\mathbf{X}; \mathcal{M}_k)}{\max p(\mathbf{X}; \mathcal{M}_0)}$$
$$= 2\ln \frac{p(\mathbf{X}; \hat{\boldsymbol{\theta}}_k)}{p(\mathbf{X}; \boldsymbol{\theta} = \mathbf{0})}$$

$\hat{\boldsymbol{\theta}}_k$ is the MLE of the unknown parameters and $u(\cdot)$ is the unit-step function. It can be shown that, $l_k(\mathbf{X})$ for each $k$ [39], [40] is

$$l_k(\mathbf{X}) = -M \ln \left(\Pi_{i=1}^k (1 - \hat{\lambda}_i^2)\right)$$

where $\hat{\lambda}_i$'s are MLEs of the circular coefficients $\lambda_i$'s. Note that $\hat{\lambda}_i$'s are the singular values of the MLE of the coherence matrix

$$\hat{\mathbf{C}}_h = \hat{\mathbf{C}}^{-\frac{1}{2}} \hat{\mathbf{P}} \hat{\mathbf{C}}^{-\frac{T}{2}}$$

where $\hat{\mathbf{C}}$ and $\hat{\mathbf{P}}$ are MLEs of $\mathbf{C}$ and $\mathbf{P}$ as follows.

$$\hat{\mathbf{C}} = \frac{1}{M} \sum_{k=1}^M (\mathbf{x}_k - \boldsymbol{\mu}_\mathrm{x})(\mathbf{x}_k - \boldsymbol{\mu}_\mathrm{x})^H$$

and

$$\hat{\mathbf{P}} = \frac{1}{M} \sum_{k=1}^M (\mathbf{x}_k - \boldsymbol{\mu}_\mathrm{x})(\mathbf{x}_k - \boldsymbol{\mu}_\mathrm{x})^T,$$

where $\boldsymbol{\mu}_\mathrm{x} = \frac{1}{M} \sum_{k=1}^M \mathbf{x}_k$ is the sample mean.

Furthermore, under hypothesis $\mathcal{H}_k$ or equivalently the model $\mathcal{M}_k$, the number of unknown parameters is $d_k = k(2N - $ $k+1$) more than that under $\mathcal{H}_0$ [5], [40]. Therefore, the Bayesian EEF rule for estimating the degree of noncircularity is:

$$\text{BEEF}_k = \left\{ -M \ln \left( \Pi_{i=1}^k (1-\hat{\lambda}_i^2) \right) \right.$$
$$\left. -d_k \left[ \ln \left( \frac{-M \ln \left( \Pi_{i=1}^k (1-\hat{\lambda}_i^2) \right)}{d_k} \right) + 1 \right] \right\}$$
$$\cdot u \left( -M \ln \left( \Pi_{i=1}^k (1-\hat{\lambda}_i^2) \right) - d_k \right),$$

where $d_k = k(2N - k + 1)$.

The estimate of the degree of noncircularity, $\hat{k}$, is the $k$ associated with $\text{BEEF}_k$ which is the maximum among all $\text{BEEF}$'s for $k = 1, \ldots, N$.

To compare the asymptotic Bayesian EEF with the MDL and AIC, we also give The MDL rule for estimating the degree of noncircularity as [9], [40]

$$\text{MDL}_k = -M \ln \left( \Pi_{i=1}^k (1-\hat{\lambda}_i^2) \right) - d_k \ln M, \quad (23)$$

MDL rule chooses the $k$ from $1, \ldots, N$ that maximizes (23).

AIC and its corrected form AICc are

$$\text{AIC}_k = -M \ln \left( \Pi_{i=1}^k (1-\hat{\lambda}_i^2) \right) - 2d_k \quad (24)$$

and

$$\text{AICc}_k = -M \ln \left( \Pi_{i=1}^k (1-\hat{\lambda}_i^2) \right) - \frac{2d_k M}{M - d_k - 1} \quad (25)$$

respectively. In addition, there are various versions of the MDL, such as gMDL [25] and nMDL [41]. However, they are derived for linear model [25], [42]. We hence do not compare the asymptotic Bayesian EEF methods with them.

### B. Numerical Simulations and Performances

Computer simulations with a similar setup as that in [40] are used to evaluate the performance of the asymptotic Bayesian EEF estimator for the degree of noncircularity and to compare performances among different model order selection rules.

In Simulation 1, for each trial we generate $M = 500$ vectors, which are drawn IID from a $N = 6$ variate complex normal distribution $\mathcal{CN}(\mathbf{0}, \mathbf{C}, \mathbf{P})$ with $\mathbf{C} = \mathbf{I}$ being an identity matrix and the pseudo-covariance matrix $\mathbf{P} = \mathbf{\Lambda}$ being a diagonal matrix with $k$ nonzero diagonal elements. The $k$ diagonal elements are in fact $k$ circularity coefficients and each of them is generated independently from the uniform distribution $U(0.05, 0.99)$ for each vector. In total, 1000 trials are run to calculate the probability of correct order $p_c$, i.e., the number of correct estimates of $\hat{k} = \text{true } k$ over the number of trials. Correct order for each $k = 1, \ldots, d$. Figure 1 shows the probability of correct order of asymptotic EEF method, MDL, AICc and AIC for different true $k$'s. As shown, the asymptotic Bayesian EEF, in general, outperforms the other methods. MDL has the tendency to favor simpler models. AIC and AICc have the tendency to favor complex models.

Simulation 2 investigates the performances of model order selection rules in a more difficult situation, i.e., the nonzero circularity coefficients on average are smaller (closer to zero) compared with those in Simulation 1. We keep $M = 500$ for this simulation but generate circularity coefficients by using a uniform distribution $U(0.05, 0.50)$ instead of the previous distribution $U(0.05, 0.99)$. This setup represents a case when the circularity coefficients have a lower average value. As shown in Figure 2 the Bayesian EEF outperforms the competing methods substantially although all the methods' performances drop correspondingly. asymptotic Bayesian EEF still outperforms other methods considerably.

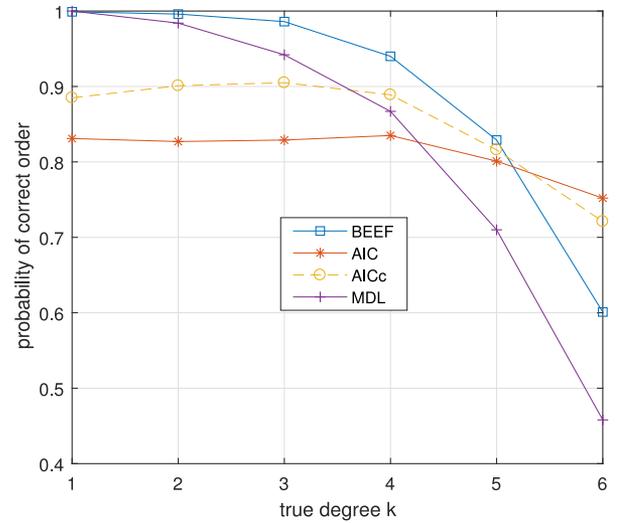

Fig. 1. Model order selection rules' performances in Simulation 1.

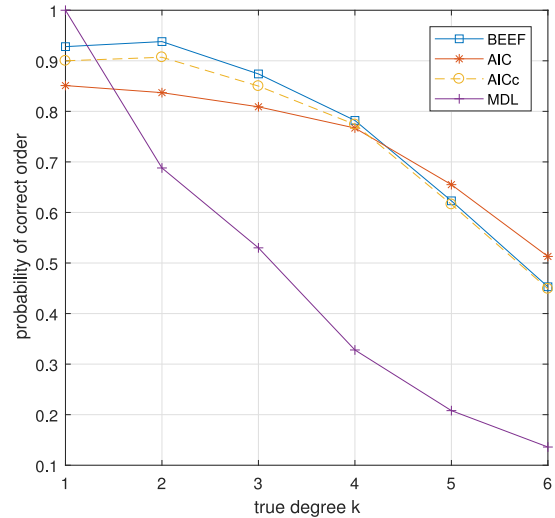

Fig. 2. Model order selection rules' performances in Simulation 2.

Simulation 3 investigates the performances of these model order selection rules with a smaller data record length. We keep generating circularity coefficients by using a uniform distribution $U(0.05, 0.99)$ as in Simulation 1. But, the number of observed vectors $M$ for this simulation is only $M = 100$. Figure 3 shows each rule's performance in this setup. Compared to Simulation 1, all methods' performance drop due to less observed data. The asymptotic Bayesian EEF outperforms its competing methods in this shorter data record situation.

Contrast to Simulation 3, in Simulation 4 the data record length $M$ for this simulation is increased to $M = 1000$. All other setup remains the same as that of Simulation 3. The results are presented in Figure 4. As shown, all methods' performances improve due to a larger data record. Bayesian EEF again has the best performance for most model order cases in this larger data record scenario.

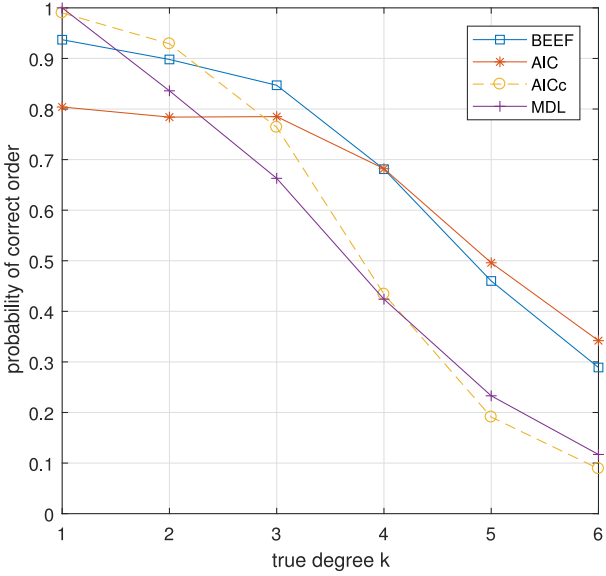

Fig. 3. Model order selection rules' performances in Simulation 3.

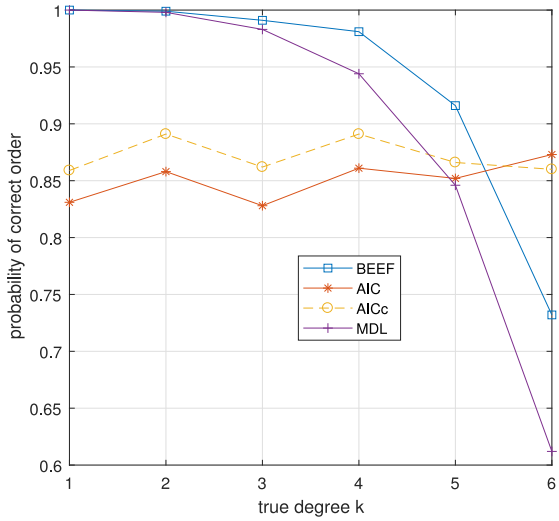

Fig. 4. Model order selection rules' performances in Simulation 4.

Next a brief analysis is given on the methods' convergence rates. The convergence rate of a model order selection rule is represented in the number of IID vectors that are needed to achieve $p_c = 1$ given certain average amplitude of circularity coefficients. We keep the mechanism of choosing the circularity coefficients as that of Simulation 1, and change the number of the vectors $M$ gradually from small number to larger ones until when the $p_c$ achieves 1 for the first time. The number of data length $M$ that allows a model order selection rule converges is called $M_c$. The larger an algorithm's $M_c$, the slower the algorithm converges. We choose a median order $k = 3$, the $M_c$ for Bayesian EEF is around 2000, while AIC and AICc will need to take about 3000 IID vectors to converge.

## VI. CONCLUSION

We have derived the Bayesian EEF, a new Bayesian model order selection rule, by using the EEF strategy in a Bayesian framework. The Bayesian EEF is shown to possess some desirable properties. To avoid introducing subjectivity in choosing parameter priors, the Bayesian EEF can utilize a vague proper prior as well as an improper non-informative prior, both of which are natural choices of non-informative priors but are usually forbidden by Bayesian model selection methods. It is also demonstrated that the EEF model order selection rule has a very intuitive penalty term as the sum of the parameter dimension and the estimated MI between the parameter and received data. This interpretation not only helps in understanding the mechanisms at work in the EEF method but also provides new insights into the open question of designing an optimal penalty term for model selection. The decomposition KLD = SNR-MI may apply to other Bayesian model selection methods and hence can probably lend new perspectives on revealing their mechanisms. Some interesting interactions and coincidences between the EEF model order selection rules derived from Bayesian and frequentist viewpoints are also explained. The derived asymptotic Bayesian EEF that uses Jeffreys' prior is applied to estimate the degree of noncircularity of complex-valued random vectors, which is essentially a model order selection problem. The numerical simulations show that the asymptotic Bayesian EEF has good performance and outperforms many other methods such as MDL, AIC and AICc. In a future work, the Bayesian EEF will be derived for the case of unknown noise variance.

## APPENDIX A
### DERIVATION OF ESTIMATED SNR AND MI TERMS

*Appendix A.1 Estimated SNR for Linear Model Case*

The estimated SNR term is found as follows for the linear model

$$\widehat{\text{SNR}} = \int \int p_{\eta_i}(\mathbf{x}, \boldsymbol{\theta}_i) \ln \frac{p_{\eta_i}(\mathbf{x}|\boldsymbol{\theta}_i)}{p_0(\mathbf{x})} d\boldsymbol{\theta}_i \, d\mathbf{x} \quad (26)$$

$$= \int_{\boldsymbol{\theta}_i} \pi'(\boldsymbol{\theta}_i) \left[ \text{KL}(p_{\eta_i}(\mathbf{x}|\boldsymbol{\theta}_i) \| p_0(\mathbf{x})) \right] d\boldsymbol{\theta}_i$$

$$= \int_{\boldsymbol{\theta}_i} \pi'(\boldsymbol{\theta}_i) \left[ \text{KL}\left(\mathcal{N}(\mathbf{H}_i \boldsymbol{\theta}_i, \sigma^2 \mathbf{I}) \| \mathcal{N}(\mathbf{0}, \sigma^2 \mathbf{I})\right) \right] d\boldsymbol{\theta}_i$$

$$= \int_{\boldsymbol{\theta}_i} \pi'(\boldsymbol{\theta}_i) \left( \frac{1}{2} \frac{\boldsymbol{\theta}_i^T \mathbf{H}_i^T \mathbf{H}_i \boldsymbol{\theta}_i}{\sigma^2} \right) d\boldsymbol{\theta}_i \quad (27)$$

$$= \int_{\boldsymbol{\theta}_i} \left[ \frac{e^{-\frac{1}{2} \boldsymbol{\theta}_i^T \left[ \frac{\eta_i}{1-\eta_i} \sigma^2 (\mathbf{H}_i^T \mathbf{H}_i)^{-1} \right]^{-1} \boldsymbol{\theta}_i}}{\sqrt{\left| 2\pi \frac{\eta_i}{1-\eta_i} \sigma^2 (\mathbf{H}_i^T \mathbf{H}_i)^{-1} \right|}} \right.$$

$$\left. \cdot \left( \frac{1}{2} \frac{\boldsymbol{\theta}_i^T \mathbf{H}_i^T \mathbf{H}_i \boldsymbol{\theta}_i}{\sigma^2} \right) \right] d\boldsymbol{\theta}_i \bigg|_{\eta_i = \hat{\eta}_i} \quad (28)$$

$$= \frac{1}{2} \frac{\mathbf{x}^T \mathbf{P}_i \mathbf{x}}{\sigma^2} - \frac{k_i}{2}$$

$$= l_{G_i} - \frac{k_i}{2} \quad (29)$$

where we have used the $\hat{\eta}_i$ in (5), treated as a constant, to replace $\eta_i$.

*Appendix A.2 Asymptotic Estimated SNR Term for General Model*

Next, for general models, we show that $\widehat{\text{SNR}} = l_{G_i} - \frac{k_i}{2}$ holds for large data record. Rewrite the $\widehat{\text{SNR}}$ term as

$$\widehat{\text{SNR}} = \int\int p_{\hat{\eta}_i}(\mathbf{x}, \boldsymbol{\theta}_i) \ln \frac{p_{\hat{\eta}_i}(\mathbf{x}|\boldsymbol{\theta}_i)}{p_0(\mathbf{x})} d\boldsymbol{\theta}_i \, d\mathbf{x}$$

$$= \int_{\mathbf{x}} p_{\hat{\eta}_i}(\mathbf{x}) \int_{\boldsymbol{\theta}_i} \pi(\boldsymbol{\theta}_i|\mathbf{x}) \left[\ln \frac{p_{\hat{\eta}_i}(\mathbf{x}|\boldsymbol{\theta}_i)}{p_0(\mathbf{x})}\right] d\boldsymbol{\theta}_i \, d\mathbf{x} \quad (30)$$

where $\pi(\boldsymbol{\theta}_i|\mathbf{x})$ is the posterior distribution of $\boldsymbol{\theta}_i$ after observing $\mathbf{x}$. For large data records we have approximately [31]

$$\pi(\boldsymbol{\theta}_i|\mathbf{x}) = \mathcal{N}(\hat{\boldsymbol{\theta}}_i, \mathbf{I}^{-1}(\hat{\boldsymbol{\theta}}_i)),$$

where $\mathbf{I}(\hat{\boldsymbol{\theta}}_i)$ is the Fisher information matrix (FIM) of $\boldsymbol{\theta}_i$ evaluated at its MLE $\hat{\boldsymbol{\theta}}_i$. And using the Laplace approximation we have

$$\int_{\boldsymbol{\theta}_i} \pi(\boldsymbol{\theta}_i|\mathbf{x}) \ln \frac{p_{\hat{\eta}_i}(\mathbf{x}|\boldsymbol{\theta}_i)}{p_0(\mathbf{x})} d\boldsymbol{\theta}_i$$

$$\approx \int_{\boldsymbol{\theta}_i} \pi(\boldsymbol{\theta}_i|\mathbf{x}) \left[ \underbrace{\ln \frac{p_{\hat{\eta}_i}(\mathbf{x}|\hat{\boldsymbol{\theta}}_i)}{p_0(\mathbf{x})}}_{l_{G_i}} \right.$$

$$\left. - \frac{1}{2}(\boldsymbol{\theta}_i - \hat{\boldsymbol{\theta}}_i)^T \mathbf{I}(\hat{\boldsymbol{\theta}}_i)(\boldsymbol{\theta}_i - \hat{\boldsymbol{\theta}}_i) \right] d\boldsymbol{\theta}_i$$

$$= l_{G_i} - \frac{k_i}{2}.$$

Therefore from (30)

$$\widehat{\text{SNR}} \approx \int p_{\hat{\eta}_i}(\mathbf{x}) \left[ l_{G_i} - \frac{k_i}{2} \right] d\mathbf{x}$$

$$= \int p(\mathbf{x}; \hat{\eta}_i) \left( T_i(\mathbf{x}) - \frac{k_i}{2} \right) d\mathbf{x}$$

$$= T_i(\mathbf{x}) - \frac{k_i}{2}$$

$$= l_{G_i} - \frac{k_i}{2} \quad (31)$$

where we have used $\left[ \int p(\mathbf{x}; \eta_i) T_i(\mathbf{x}) d\mathbf{x} \right]\big|_{\eta_i = \hat{\eta}_i} = T_i(\mathbf{x})$ in (7).

*Appendix A.3 Estimated MI Term*

This subsection derives the estimated MI term as follows.

$$\widehat{\text{MI}}$$

$$= \int\int p_{\hat{\eta}_i}(\mathbf{x}, \boldsymbol{\theta}_i) \ln \frac{p_{\hat{\eta}_i}(\mathbf{x}|\boldsymbol{\theta}_i)}{p_{\hat{\eta}_i}(\mathbf{x})} d\mathbf{x} d\boldsymbol{\theta}_i \quad (32)$$

$$= \int_{\boldsymbol{\theta}_i} \pi'(\boldsymbol{\theta}_i) \int_{\mathbf{x}} p_{\hat{\eta}_i}(\mathbf{x}|\boldsymbol{\theta}_i) \ln \frac{p_{\hat{\eta}_i}(\mathbf{x}|\boldsymbol{\theta}_i)}{p_{\hat{\eta}_i}(\mathbf{x})} d\mathbf{x} d\boldsymbol{\theta}_i$$

$$= \int_{\boldsymbol{\theta}_i} \pi'(\boldsymbol{\theta}_i) \text{KL}\left( p_{\hat{\eta}_i}(\mathbf{x}|\boldsymbol{\theta}_i) || p_{\hat{\eta}_i}(\mathbf{x}) \right) d\boldsymbol{\theta}_i$$

$$= \int_{\boldsymbol{\theta}_i} \pi'(\boldsymbol{\theta}_i)$$

$$\cdot \text{KL}\left( \mathcal{N}(\mathbf{H}_i \boldsymbol{\theta}_i, \sigma^2 \mathbf{I}) || \mathcal{N}\left(\mathbf{0}, \sigma^2 \mathbf{I} + \frac{\hat{\eta}_i}{1-\hat{\eta}_i}\sigma^2 \mathbf{P}_i\right) \right) d\boldsymbol{\theta}_i$$

$$= \int_{\boldsymbol{\theta}_i} \pi'(\boldsymbol{\theta}_i) \left[ \frac{1}{2} \ln \frac{|\sigma^2 \mathbf{I} + \frac{\hat{\eta}_i}{1-\hat{\eta}_i}\sigma^2 \mathbf{P}_\mathbf{H}|}{|\sigma^2 \mathbf{I}|} \right.$$

$$+ \frac{1}{2} \text{tr}\left( \sigma^2 \left(\sigma^2 \mathbf{I} + \frac{\hat{\eta}_i}{1-\hat{\eta}_i}\sigma^2 \mathbf{P}_\mathbf{H}\right)^{-1} - \mathbf{I} \right)$$

$$\left. + \frac{1}{2}(\mathbf{H}\boldsymbol{\theta})^T \left(\sigma^2 \mathbf{I} + \frac{\hat{\eta}_i}{1-\hat{\eta}_i}\sigma^2 \mathbf{P}_\mathbf{H}\right)^{-1} \mathbf{H}\boldsymbol{\theta} \right] d\boldsymbol{\theta}_i$$

$$= \frac{1}{2} \ln \frac{|\sigma^2 \mathbf{I} + \frac{\hat{\eta}_i}{1-\hat{\eta}_i}\sigma^2 \mathbf{P}_\mathbf{H}|}{|\sigma^2 \mathbf{I}|}$$

$$+ \frac{1}{2} \text{tr}\left( \sigma^2 \left(\sigma^2 \mathbf{I} + \frac{\hat{\eta}_i}{1-\hat{\eta}_i}\sigma^2 \mathbf{P}_\mathbf{H}\right)^{-1} - \mathbf{I} \right)$$

$$+ \int_{\boldsymbol{\theta}_i} \left[ \pi'(\boldsymbol{\theta}_i) \frac{1}{2}(\mathbf{H}\boldsymbol{\theta})^T \left(\sigma^2 \mathbf{I} + \frac{\hat{\eta}_i}{1-\hat{\eta}_i}\sigma^2 \mathbf{P}_\mathbf{H}\right)^{-1} \mathbf{H}\boldsymbol{\theta} \right] d\boldsymbol{\theta}_i$$

$$= \frac{1}{2} \ln \frac{\left|\sigma^2 \mathbf{I} + \frac{\hat{\eta}_i}{1-\hat{\eta}_i}\sigma^2 \mathbf{P}_i\right|}{|\sigma^2 \mathbf{I}|}$$

$$= \frac{k_i}{2} \ln\left(\frac{1}{1-\hat{\eta}_i}\right) \quad (33)$$

$$= \frac{k_i}{2} \ln\left(\frac{\mathbf{x}^T \mathbf{P}_i \mathbf{x}}{k_i \sigma^2}\right) \quad (34)$$

$$= \frac{k_i}{2} \ln \frac{2 l_{G_i}}{k_i} \quad (35)$$

ACKNOWLEDGMENT

The authors would like to thank the Associate Editor and the anonymous reviewers for their constructive comments toward the improvement of this paper.

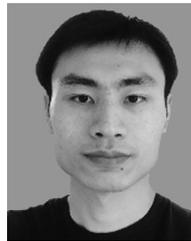

**Zhenghan Zhu** (M'17) received the B.S. and M.S. degrees from Shanghai JiaoTong University, Shanghai, China, in 2008 and 2011, respectively, and the Ph.D. degree from the University of Rhode Island, Kingston, RI, USA, in 2017, all in electrical engineering.

In 2011, he was an Application Engineer with Teradyne (Shanghai). He is currently a Radar Signal Processing Engineer with Delphi. His research interests include estimation and detection theory, information-theoretic signal processing, statistical machine learning, automotive radar signal processing, and autonomous driving algorithm design.

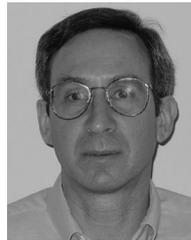

**Steven Kay** (LF'89) was born in Newark, NJ, USA, on April 5, 1951. He received the B.E. degree from Stevens Institute of Technology, Hoboken, NJ, USA, the M.S. degree from Columbia University, New York, NY, USA, and the Ph.D. degree from Georgia Institute of Technology, Atlanta, GA, USA, in 1972, 1973, and 1980, respectively, all in electrical engineering.

From 1972 to 1975, he was with Bell Laboratories, Holmdel, NJ, USA, where he was involved with transmission planning for speech communications and simulation and subjective testing of speech processing algorithms. From 1975 to 1977, he attended Georgia Institute of Technology to study communication theory and digital signal processing. From 1977 to 1980, he was with the Submarine Signal Division, Portsmouth, RI, USA, where he engaged in research on autoregressive spectral estimation and the design of sonar systems. He is currently a Professor in electrical engineering with the University of Rhode Island, Kingston, RI, USA, and a Consultant to industry and the Navy. He has authored or coauthored numerous papers and is a contributor to several edited books. He is the author of the textbooks *Modern Spectral Estimation* (Prentice-Hall, 1988), *Fundamentals of Statistical Signal Processing, Vol. I: Estimation Theory* (Prentice-Hall, 1993), *Fundamentals of Statistical Signal Processing, Vol. II: Detection Theory* (Prentice-Hall, 1998), and *Intuitive Probability and Random Processes Using MATLAB* (Springer, 2006). His research interests include spectrum analysis, detection and estimation theory, and statistical signal processing.

Dr. Kay is a Member of Tau Beta Pi and Sigma Xi. He was on the IEEE Acoustics, Speech, and Signal Processing Committee on Spectral Estimation and Modeling and on several IEEE Oceans committees. He was an Associate Editor for the IEEE SIGNAL PROCESSING LETTERS and was a Distinguished Lecturer for the IEEE Signal Processing Society. Recently, he has been included in a list of the 250 most cited researchers in the world in engineering. He was the recipient of the IEEE Signal Processing Society Education Award.